\def\year{2019}\relax
\documentclass[letterpaper]{article} 
\usepackage{aaai19}  
\usepackage{times}  
\usepackage{helvet}  
\usepackage{courier}  
\usepackage{url}  
\usepackage{graphicx}  
\usepackage[round]{natbib}
\usepackage{amssymb}
\usepackage{xcolor}
\usepackage{amsmath}
\usepackage{textcomp}
\usepackage{siunitx}
\usepackage{graphicx}
\usepackage{caption}
\usepackage{booktabs}
\usepackage{multirow}
\usepackage{tabularx}
\usepackage{algorithmic}
\usepackage{algorithm}
\usepackage{graphicx,color}
\usepackage{graphicx,subfigure}
\begin{document}
%
\title{Autonomous Extraction of a Hierarchical Structure of Tasks in Reinforcement Learning, A Sequential Associate Rule Mining Approach}
\author{Behzad Ghazanfari$^{\dagger}$, Fatemeh Afghah$^{\dagger}$, Matthew E. Taylor$^{\ddagger}$\\
beghazanfari@gmail.com, Fatemeh.Afghah@nau.edu\\
$^{\dagger}$ School of Informatics, Computing, and Cyber Security, Northern Arizona University \\
taylorm@eecs.wsu.edu\\
$^{\ddagger}$ School of Electrical Engineering and Computer Science, Washington State University\\
}
\maketitle
\begin{abstract}

Reinforcement learning (RL) techniques, while often powerful, can suffer from slow learning speeds, particularly in high dimensional spaces. Decomposition of tasks into a hierarchical structure holds the potential to significantly speed up learning, generalization, and transfer learning. However, the current task decomposition techniques often rely on high-level knowledge provided by an expert (e.g. using dynamic Bayesian networks) to extract a hierarchical task structure; which is not necessarily available in autonomous systems.
In this paper, we propose a novel method based on \textit{Sequential Association Rule Mining} that can extract \textit{Hierarchical Structure of Tasks in Reinforcement Learning} (\textit{SARM-HSTRL}) in an autonomous manner for both Markov decision processes (MDPs) and factored MDPs. 
The proposed method leverages association rule mining to discover the causal and temporal relationships among states in different trajectories, and extracts a task hierarchy that captures these relationships among sub-goals as termination conditions of different sub-tasks. We prove that the extracted hierarchical policy offers a hierarchically optimal policy in MDPs and factored MDPs. It should be noted that \textit{SARM-HSTRL} extracts this hierarchical optimal policy without having dynamic Bayesian networks in scenarios with a single task trajectory and also with multiple tasks' trajectories. Furthermore, it has been theoretically and empirically shown that the extracted hierarchical task structure is consistent with trajectories and provides the most efficient, reliable, and compact structure under appropriate assumptions. The numerical results compare  the performance of the proposed \textit{SARM-HSTRL} method with conventional HRL algorithms in terms of the accuracy in detecting the sub-goals, the validity of the extracted hierarchies, and the speed of learning in several testbeds.

\end{abstract}

\section{Introduction}
\noindent  Reinforcement learning is known as a commonly used approach for planning and sequential decision making in artificial intelligent (AI) systems, where the agents gradually learn and optimize their actions from delayed rewards through a trial-and-error mechanism. However, one of the main challenges of RL approaches is scalability to high-dimensional state spaces \citep{barto2003recent}. Hierarchical reinforcement learning (HRL) methods are known to reduce the computational complexity of RL approaches by temporal and state abstraction in the form of decomposing the learning problem to a hierarchy of several sub-problems. Sub-goals refer to the local target states that not only provide easy access or high reinforcement gradients, but also must be visited frequently \citep{mcgovern2002autonomous, stolle2004automated}. These sub-goals can help an agent to accelerate the learning process, particularly in high dimensional spaces. 
In \citep{dietterich2000hierarchical}, a HRL decomposition method called MAXQ is proposed based on the assumption of having an expert with the knowledge of sub-goals to provide a correct hierarchy, however such assumption can restrict the application of this method in autonomous systems where a limited expert's understanding is available \citep{taylor2009transfer}.  

In the absence of an expert, several HRL techniques have been reported for task decomposition, in which a number of sub-goals that are correlated with the successful policies are utilized as the required states to decompose the learning task \citep{digney1998learning, mcgovern2002autonomous, stolle2004automated}. However, extracting these states in an autonomous manner is still a challenging problem \citep{chiu2011subgoal}. More importantly, in the majority of existing HRL methods, the potential hidden correlations among these sub-goals to achieve the ultimate goal have been overlooked. In general, the current HRL methods with autonomous task decomposition capability can be divided into two groups depending on their domains. 

\noindent \textbf{HRL methods in MDPs:} The HRL methods based on extracting the sub-goals \citep{mcgovern2002autonomous, stolle2004automated}, or the ones based on bottlenecks extraction \citep{mannor2004dynamic,csimcsek2004using, csimcsek2009skill} can only extract a flat hierarchy, (i.e., one level) which means that these methods only find the sub-goals or the bottlenecks, rather than a hierarchical structure of those. Since these methods often use the paths or the sub-graphs of the agent, or the shortest paths among the nodes of a graph to calculate their required measures such as betweenness \citep{csimcsek2009skill}, their performance considerably degrades in scenarios with a large state space, or when the number of actions to reach the goal-states increases. They also usually require prior knowledge about the measures that helps to partition state space to parts that are connected densely inside, but sparsely to each other \citep{ghazanfari2016extracting}. 

\noindent \textbf{HRL methods in Factored MDPs (FMDPs):} Some of the current HRL methods based on extracting the task-dependent hierarchy in FMDPs include HEX-Q \citep{hengst2003discovering}, VISA \citep{jonsson2006causal}, and HI-MAT \citep{mehta2008automatic, mehta2011automatic}. Since there are implicit structure representations of the problems among states variable in FMDPs, dynamic Bayesian networks (DBNs) as high-level sources of pre-knowledge are often utilized to decompose the tasks in such processes noting their capability to extract the impact of each action on state variables. HI-MAT and VISA algorithms rely on availability of DBNs for each action \citep{jonsson2006causal, mehta2008automatic, mehta2011automatic}.
Since VISA considers the impacts of all actions regardless of the domain, it can create unnecessary branches in the extracted hierarchy or unnecessary subtasks. Thus, it may result in an ``exponentially sized hierarchy'' that limits its application in some domains \citep{mehta2008automatic, mehta2011automatic}.
To address this problem, HI-MAT was proposed to remove such unsuccessful and redundant actions cycles. This method leverages a single and carefully constructed trajectory to construct a MAXQ hierarchy. It is shown that the constructed hierarchy is compact and comparable to manually engineered ones.
However, the main disadvantage of both these methods is that utilizing DBNs require high-level knowledge that should be provided by an expert or needs to be extracted via a large number of computations \citep{wynkoop2008learning}. Among these HRL methods proposed for factored MDPs, HEX-Q is the only one that does not rely on DBNs. 
However, this method is not capable of identifying the relations among the states variables that can potentially results in divergence of the learning process \citep{mehta2008automatic}. 
\cite{bacon2017option} used a policy gradient method to create temporally extended actions instead of extracting the sub-goals. However, this method can only handle one task at the time and needs to know the number of options in advance; therefore, it may have limited application in multi-task RL, or in cases with a large number of subtasks.

The key contribution of this work is to propose a HRL method based on the idea of sequential association rule mining \textit{(SARM)} that extracts a hierarchical knowledge from the hidden correlations among the extracted sub-goals and use this knowledge to decompose the tasks to multiple sub-tasks. 
Conventional subgoal extraction methods that can work in MDPs, do not extract a hierarchal task structure. The few existing hierarchal structure extractor methods in RL including HEX-Q, HI-MAT, and VISA only work in FMDPs. More importantly, HI-MAT and VISA rely on DBNs knowledge, which is a high-level supplementary knowledge provided by human experts. HI-MAT, the most recent HRL approach in the literature has the following limitations: 1) requiring DBNs knowledge, 2) cannot work based on several trajectories that are a typical situation in RL, 3) cannot support funnel property of a subtask, and 4) cannot be applied in MDPs. However, our proposed \textit{SARM-HSTRL} method extracts a hierarchical optimum policy task structure for both MDPs and FMDPs, while it does not rely on DBNs as a pre-knowledge structure provided by human experts in FMDPs. To the best of our knowledge, \textit{SARM-HSTRL} is also the first method that can extract the hierarchical optimum policy task structure of multiple policies. 

\section{An Overview on Association Rule Mining} \label{Sec:background}
Association rule mining (ARM) methods use a combination of two key measures of \textit{support} and \textit{confidence} in a proven efficient extraction strategy to obtain and evaluate the most efficient and reliable relationships among the variables in a dataset. ARM has been applied in bioinformatics to discover the patterns in datasets that are statistically important \citep{bebek2007pathfinder}, or in retail stores to find the items that are commonly being sold together among millions of transactions \citep{lin2002efficient,tan2006introduction}.

An ARM problem is defined by a pair of $\langle ITEMSET,\ Transaction \rangle$, where \textit{ITEMSET} = $\{i_{1}, \dots ,i_{g}\}$ is the set of all items and \textit{Transaction} = $\{\Omega_{1}, \dots ,\Omega_{N}\}$ is the set of all transactions. Each transaction is a subset of items of \textit{ITEMSET}. 
 The relationship among the items in the transaction set can be defined by an \textit{association rule}. An association rule is expressed in the form of $A\rightarrow B$, where $A$ and $B$ are disjoint sets of items; $A\cap B$ = $\emptyset$. The frequency of the occurrence of $A$ and $B$ together in a \textit{Transaction} is defined as a \textit{key factor}, also known as \textit{support} of the association rule. The frequency of occurrence of $A$ and $B$, relative to the frequency of the occurrence of $A$, is known as \textit{confidence}. The definition of support and confidence are as follows \citep{tan2006introduction}:
\setlength{\belowdisplayskip}{1pt}
\setlength{\abovedisplayskip}{1pt}
\setlength{\abovedisplayskip}{1pt}
\[
\textit{support} (A \rightarrow B) = \frac{\sigma(A \cup B)}{N} \]

\[\textit{confidence} (A \rightarrow B) = \frac{\sigma(A \cup B)}{\sigma (A)}
\]
\setlength{\belowdisplayskip}{1pt}
\setlength{\abovedisplayskip}{1pt}
\noindent where $\sigma(\cdot)$ is the number of observed transactions including the elements inside of the parenthesis, and $N$ is the total number of transactions. The support factor is often used as a measure to disregard the items that do not occur together so frequently relative to $N$, and confidence can express the reliability of the extracted rule. The corresponding thresholds for support and confidence, known as \textit{minsup} and \textit{minconf}, respectively, can be used to extract important rules \citep{tan2006introduction}. ARM algorithms typically consist of two parts: 
1) Frequent Itemset Generation: All of the itemsets that satisfy the \textit{minsup} condition are extracted, i.e., frequent item sets.
2) Rule Generation: Building upon the outputs of the Frequent Itemset Generation, this step calculates the confidence of the obtained frequent itemsets and checks their eligibility by comparing their confidences with minconf threshold. The frequent pattern growth (FP-growth) algorithm has been proposed for Frequent Itemset Generation by constructing a compact data structure, called a FP-tree. The confidence value is calculated for each of the rules and evaluated based on minconf. This algorithm outperforms the majority of frequent patterns extraction algorithms in large datasets; for the analysis of time complexity and more details about FP-growth algorithm see \cite{kosters2003complexity, tan2006introduction}.

\section{\textit{Proposed SARM-HSTRL Algorithm}} \label{SRAM}
Here, we propose an algorithm to extract a hierarchical structure of tasks in RL named \textit{SARM-HSTRL} that works in both MDPs and FMDPs. To the best of our knowledge, despite all of the existing HRL methods in MDPs, \textit{SARM-HSTRL} extracts a hierarchical abstraction, not a flat abstraction, in MDPs. 
In continue, we define some of the terms used throughout the paper. An overview of MDPs and FMDPs, and additional details on notations and definitions are presented in the \textit{Background} subsection of the ``supplementary material" section in the end of document.

\textbf{ Definition 1:} To assign a unique representation to a set of multiple state variables in FMDPs, here we define a reversible coder-decoder operation. A map function, $MF$, as a coder maps the state variables in FMDPs to one variable. $MF^{-1}$ as a decoder is the reverse process of retrieving the FMDPs' state variables from just that single value, $L$, as described in Algorithm \ref{algkeinvers}. It means \textit{MF}, $ MF(x_{1},x_{2},\dots,x_{n})=R_{x_{1}} + \sum_{i=2}^{n}R_{x_{i}} \prod_{j=1}^{i-1}numD_{j}$, should be a surjective, injective, and invertible  function, where $D_{i}$ refers to set of possible values for each state variable, $numD_{i}$ denotes the number of possible values in $D_{i}$, and $R_{x_{i}}$ shows the index of $x_{i}$ in $numD_{i}$. Therefore,  $\prod_{i=1}^nnumD_{i}$ is the total possible number for $X$. 

\textbf{Definition 2:} Transitions are considered \textit{unpredictable} when they lead to entering or leaving subgoals. The region and the boundaries among states' clusters that have unpredictable transitions are considered as \textit{exits} and defined by a state action pair $G_{i}=(s^{T_{i}}, a)$ when taking action $a$, as a primitive action, from state $s^{T_{i}}$, as a subgoal, leads to the resultant state that is a goal state to complete subtask $T_i$ \citep{hengst2003discovering}. This concept has been further explained with an example in the ``supplementary material".

\textbf{Definition 3:} A task hierarchy $H$ is generally shown as a tree, or a directed acyclic graph, $\langle T, E\rangle$, in which the root as the main task, $T_{0}$, is decomposed to other subtasks $T_{1},\dots,T_{n}$ and the edges, $E$ represent the relation among them.  
A subtask, $T_{i}$, is a semi-MDP (SMDP) that is shown by $\langle X_{i},  S_{i}, G_{i}, C_{i}, \rangle$ \citep{mehta2011automatic}, where $X_{i}$ is the set of variables that their corresponding values change during performing the subtask, $S_{i}$ denotes the set of admissible states of $T_{i}$, $G_{i}$ shows the exits of corresponding subtasks as termination conditions of $T_{i}$, and $C_{i}$ is the set of child tasks of $T_{i}$. Child tasks, $C_{i}$, can be formed based on different HRL frameworks such as MAXQ or option.

In the task hierarchy graph, leaf nodes correspond to subtasks that interact with the environment directly by applying primitive actions, $A$, to states, $S_{i}$. Other nodes of the task hierarchy include subtasks as abstract states and their corresponding local policies as abstract actions.
We should note that the subtasks are defined over the extracted regions as the policies that lead to leaving these regions via \textit{exits}. These definitions guarantee that no action can lead to leaving a subtask except via its \textit{exits}. Each subtask similar to a region includes a set of states, actions, Markov transitions, and reward functions. 

The proposed \textit{SARM-HSTRL} decomposes the tasks into multiple subtasks by extracting the subgoals as subtask's termination, \textit{exits}, are the task graph's nodes and their relations, in the form of association rules, are the edges of the graph. The state space is partitioned recursively in a top-down manner, and the state abstraction and corresponding options exit, temporal abstraction, are formed for these partitions. The state abstraction and temporal abstraction can limit the policy search space that leads to increasing the speed of learning. 

\scalebox{1}{
\begin{minipage}[b]{7cm}
		\vspace{0pt}  
		\begin{algorithm}[H]
       
        \caption{$MF^{-1}(L)$} \label{algkeinvers}
			\begin{algorithmic}
            	
                \FOR{ $i=n:1$ } \STATE $TEMP=\prod_{j=1}^{i-1}numD_{j}$
                	\STATE $Rx_{i}=L/TEMP$, $L=mod(L,TEMP)$
                    \IF { $L==0$}
                  \STATE $Rx_{i}=Rx_{i}-1, L=TEMP$
                  \ENDIF
                \ENDFOR
                 
		\end{algorithmic}
       \end{algorithm}
\end{minipage}%
}

The proposed $\textit{SARM-HSTRL}$ is composed of two phases (see Algorithm \ref{main}).
In the first phase, several association rules are extracted using an \textit{SARM} approach following the two steps of i) Frequent Itemsets Generation, and ii) Rule Generation procedure. Then, the proposed $HST-construction$ method converts these association rules to a hierarchical structure tree\footnote{A example of applying \textit{SARM-HSTRL} on a testbed along with a detailed description of the \textit{SARM-HSTRL} process notations is presented in the ``supplementary material" section.}. 

\begin{minipage}[b]{8cm}
		\vspace{0pt}
		
		\begin{algorithm}[H]
			\caption{\textit{SARM-HSTRL}}\label{main}
			\begin{algorithmic}
			\STATE \textbf{Input}: Transition, \textit{minsup}, \textit{minconf}	
			\STATE \textbf{Output}: \textit{HST}
			
			\STATE \textbf{1.} Frequent Itemset =  FP-growth (Transition, \textit{minsup}) 
			\STATE \textbf{2.} Association Rules = Rule Generation (Frequent Itemset, \textit{minconf})
			
			\STATE \textbf{3.} HST-construction (Association Rules) // See Algorithm \ref{hstalgorithm} 
    	\end{algorithmic}
		\end{algorithm}
	\end{minipage}

In the proposed \textit{SARM-HSTRL}, each trajectory of visited states, $\Omega_{k}$=$\{s_{1}, \dots, s_{h}\}$, is considered as a transaction member of the \textit{Transaction} set, in which $h$ shows the number of states in that transaction. In FMDPs, the proposed function $MF$ (see Definition 1) is used to map multivariate state variable to a univariate state variable. All visited states in successful trajectories\footnote{A successful trajectory is defined as a trajectory of states that leads to the goal reward \citep{mehta2008automatic}.} are stored in the \textit{ITEMSET}. $T_{i}$ are defined based on sub-goal states. These sub-goals, as exit states, are defined as the states that are frequently visited in successful trajectories (i.e., the trajectories where the agent reaches a goal state). In other words, the problem of finding the sub-goals and the relations among them can be seen as extracting association rules such as $ \{s_{d}, \dots, s_{g} \rightarrow s_{h} \}$, where  $ \{s_{d}, \dots, s_{g}, s_{h} \}$ are sub-goal states. It should be noted that there often exists a set of some key subgoals that are common among different tasks, and the proposed \textit{SARM-HSTRL} method can extract such key subgoals by processing a set of trajectories of tasks with random start and goal states. 

Here, we use the FP-growth algorithm to perform the first step of \textit{SARM} called Frequent Itemset Generation. If the \textit{minsup} is set to its maximum possible value (i.e., one), the sub-goals must be visited in each trajectory of each transaction. If we set a very small value to the \textit{minsup}, the performance of FP-growth will be degraded as \textit{SARM-HSTRL} may provide some false-positive itemsets for the evaluation of Rule Generation. Hence, we face a trade-off in selecting reasonable values for \textit{minsup} and \textit{minconf}. On one hand, these values should be small enough to capture different sub-goals and relations in RL domains with multiple types of successful trajectories. On the other hand, if the \textit{minsup} and \textit{minconf} are set to very small values, the extracted hierarchical structure would extract some unnecessary sub-goals and relations. The proper range of these parameters can be set based on the number of trajectories of encountered tasks. 

\begin{algorithm} 
\caption{HST-construction : Construct a tree, $T$, with one node that is the root node, $R$. } 
\label{hstalgorithm}
\scalebox{1}{
\begin{minipage}{1\linewidth}
	\begin{algorithmic}[1]
		\STATE \textbf{Input}$:$ \textit{AR-set} is the set of association rules. \textit{AR-set} = $\{AR_{1}, \ldots ,AR_{NumRules}\}$
		
		\STATE \textbf{Output}$:$ \textit{HST}
        
        \STATE $num$ : the number of children of the Parent-Node; $PN_{t}$ : the $t_{th}$ child of the Parent-Node\ 
		
		\FOR { $i=1:NumRules$  } 
		\STATE Parent-Node=$R$\
		
		\FOR {$j=1:Len_{i}$} 
		\STATE $t=1$\ , $FlagM=0$\
		\REPEAT 
		\IF {$AR_{ij}==PN_{t}$}
		\STATE Parent-Node=$PN_{t}$; $FlagM=1$
		\ENDIF
		\STATE ${t=t+1}$
		\UNTIL{$t<=num$ and $FlagM==0$}	
		
		\IF{$FlagM==0$}
		\STATE create a new child Node in the Parent-Node: $PN_{num+1}=AR_{ij} $
		\STATE Parent-Node=$PN_{num+1}$
		\ENDIF				
		\ENDFOR
		\ENDFOR

	\end{algorithmic}
\end{minipage}
}
\end{algorithm}

\setlength{\belowdisplayskip}{0.1pt}
\setlength{\abovedisplayskip}{0.1pt}
    
Next, the Rule Generation procedure is performed on the extracted frequent itemsets as the output of FP-growth algorithm. 
Recall that a confidence value is the conditional probability of the occurrence of a consequent of a certain rule when its premise has been seen, and are calculated using \textit{minconf} thresholds. The confidence value of each association rule can be used as a priority score to choose among corresponding temporally extended actions of association rules.

Besides extracting a set of sub-goals as the association rules, \textit{SARM-HSTRL} also extracts different possible sequences of these sub-goals for HST construction in a sequential association rule mining procedure.
The value of $ t$, time of each sub-goal in each trajectory, can be compared to create a sequence of observed sub-goals.
Each sequence shows the relationship among the sub-goals in a flat manner of one association rule. For instance, is there are two trajectories of four sub-goals ${a, b, c \rightarrow d}$ and ${b, a, c \rightarrow d}$, the $t$'s values of $a$ and $b$ are $\{1 , 2\}$ and $\{2 , 1\}$, respectively, in the trajectories. If the frequencies of those orders are the same, it means that the order of visiting $a$ and $b$ is not important to achieve the consequent subgoal although each sequence could have different probability values. 

Algorithm \ref{hstalgorithm} describes the HST-construction method that makes the hierarchical structure of tasks. The HST helps an agent to choose the correct sub-tasks. Each association rule $AR_{i}$ can be shown in the form of $AR_{i}= s_{ti}, \ldots, s_{(t+n)i} \rightarrow s_{(t+n+1)i}$, where $\{ s_{ti}, \ldots, s_{(t+n)i} \}$ denotes a sequence of sub-goals of the $AR_{i}$. In this algorithm, $Len_{i}$ denotes the number of items in $AR_{i}$. The number of elements of the premise of the $AR_{i}$ is $n+1$, and the number of elements of the consequence of each $AR$ is 1; thus, the $Len_{i}$ is $n+2$.  $AR_{i,j}$ is the $j$th element from the end of $AR_{i,j}$. For example, $AR_{i,2}$ is $s_{t+n}$ and $AR_{i,len_{i}}$ is $s_{t}$. $NumRules$ is the number of association rules.

\begin{figure}
	\centering 
    \subfigure[]{\label{fig:HST_test1_hi}\includegraphics[width=25mm, height=1.2in]{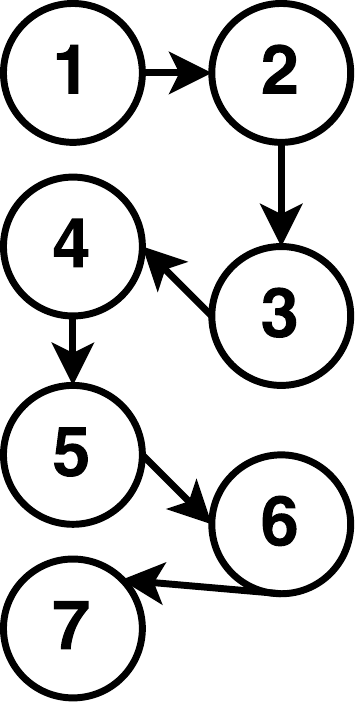}}
    \subfigure[]{\label{fig:HST_test2_hi}\includegraphics[width=50mm, height=1.2in]{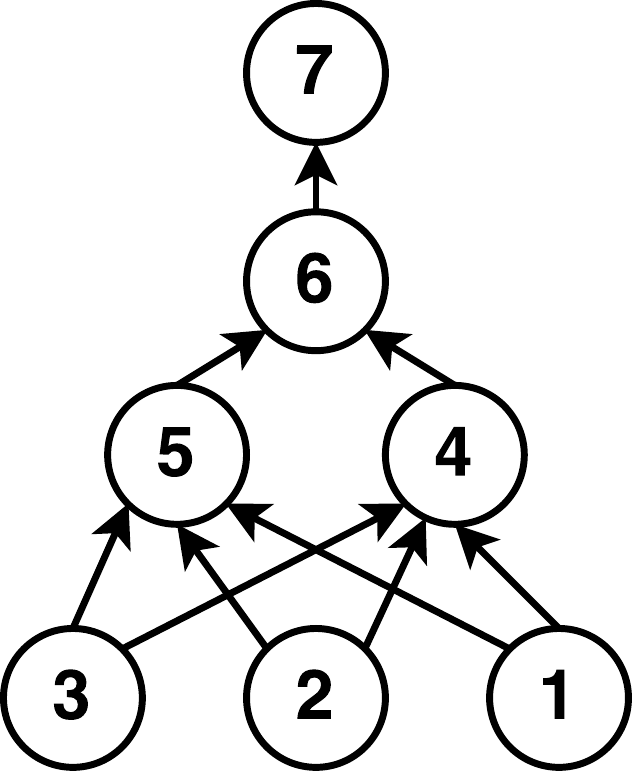}}	
\subfigure[]{\label{fig:testbed1}\includegraphics[width=80mm, height=1.7in]	{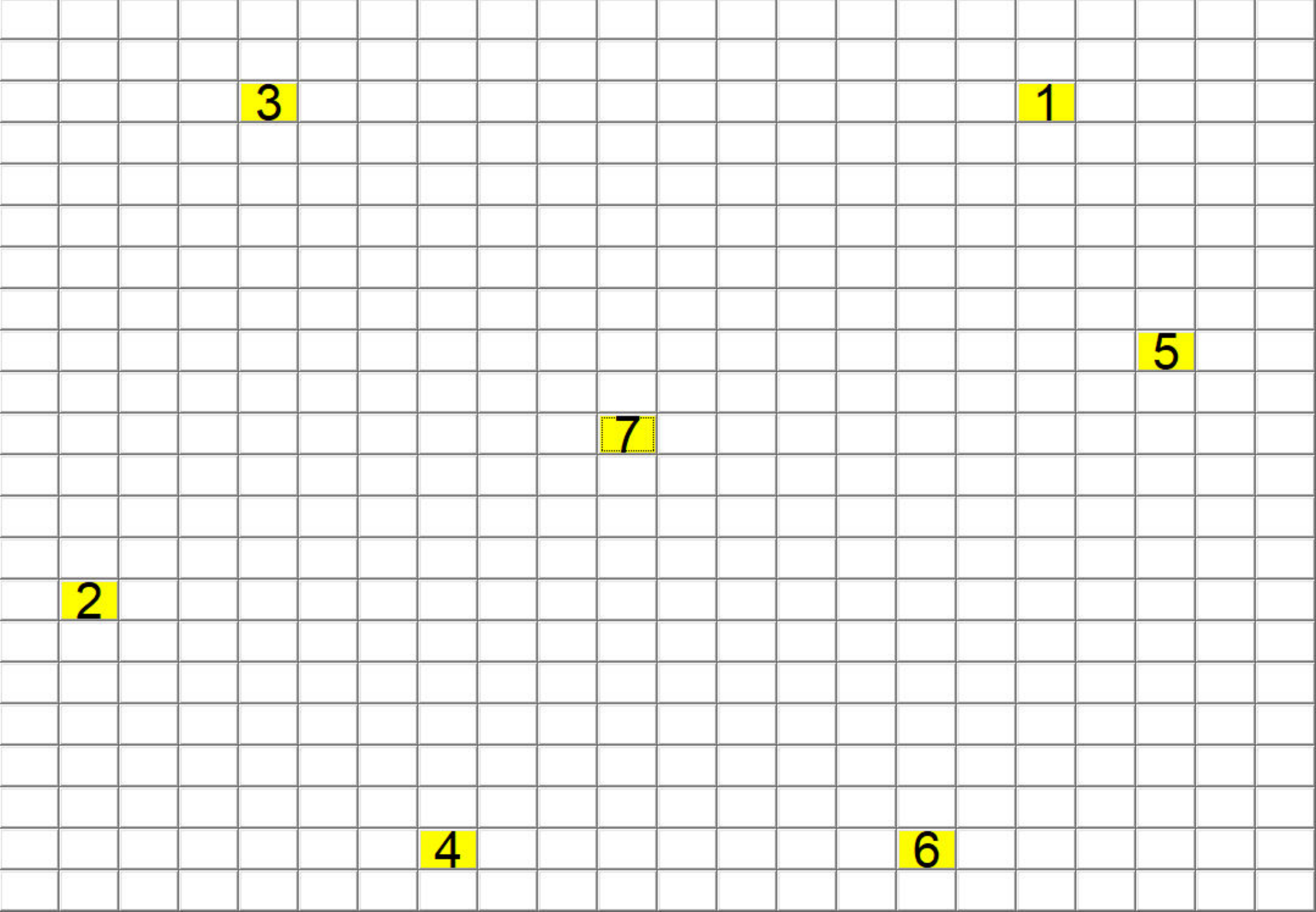}}
\vspace{-10pt}
        \caption{ \small{(a) The first task hierarchy of the first testbed, Figure \ref{fig:testbed1}, for  experiment 1. (b) The second task hierarchy of the first testbed, Figure \ref{fig:testbed1}, for experiment 2. (c) The first testbed: the size of the maze is $22\times22$ and it has 7 sub-goals. The subgoals are colored with yellow.}}
\vspace{-15pt}
\end{figure}
\setlength{\belowdisplayskip}{0.1pt}
\setlength{\abovedisplayskip}{0.1pt}

\section{Theoretical Analysis} \label{theory}
\noindent In this section, we provide a theoretical analysis to study the properties of the extracted hierarchical structure of the tasks using the proposed \textit{SARM-HSTRL} method. The problem of extracting a hierarchical structure in RL can be considered as a hierarchical credit assignment problem of MAXQ, and the proposed \textit{SARM-HSTRL} provides a solution to automatically perform such extraction in model free MDPs and FMDPs. Since the required convergence conditions of \textit{SARM-HSTRL} are different in MDPs and FMDPs, its performance has been evaluated in each domain separately.\\
%

 \textbf{Theorem 1}: The proposed \textit{SARM-HSTRL} converges to a hierarchical optimum policy in model-free MDPs. A hierarchical policy is defined as an assignment of a local policy to each subtask. A hierarchical optimum policy is a hierarchical policy that makes the best accumulated reward \citep{mehta2011automatic}.\\ 

\textbf{Proof}: Using Definition 2 in \cite{hengst2003discovering}, the proof that the extracted HST leads to a hierarchal optimal policy is straightforward if we show that only one variable of X changes as a result of each action; because then the Q function of such hierarchy can be recursively expanded and mapped to a Q function of a flat MDP. In our proposed method in MDPs, each node of HST corresponds to a sub-MDP and based on Definition 1,  there exists only one variable in our state variable, $X$. Therefore, as proved for HEX-Q algorithm, the proposed \textit{SARM-HSTRL} converges to a hierarchical optimum policy. Hierarchical execution can be applied by using a decomposed value function since the proposed method similar to MAXQ breaks down the MDP to interlinked sub-MDPs directly. Q function for each node, exit, as a sub-MDP of the tree is defined recursively as follows:
\[Q^{*}_{T_{i}}(s^{T_{i}},a) = \sum_{s^{'}}P_{s^{T_{i}}s^{'}}^\alpha [R_{s^{T_{i}}s^{'}}^\alpha + V_{T_{i}}^{*}(s^{'})]\]
where $s^{'}$ is the hierarchical next state. $Q^{*}_{T_{i}}(s^{T_{i}},a)$ shows the expected value of node $T_{i}$ after performing (abstract) action $a$ in (abstract) state $s^{T_{i}}$ and in the continue pursuing the optimal hierarchical policy. $V^{*}_{T_{i}}(s)$ is the decomposition of optimal hierarchical value function that is calculated recursively as follow:

\[V^{*}_{T_{i}}(s) =\max_{a} [V^{*}_{C_{i}(a)}(s)+Q^{*}_{T_{i}}(s^{T_{i}},a)] \]

where $V^{*}_{C_{i}(a)}$ shows the child of $T_{i}$ implementing action $a$.  
$\Box$ \\

In continue, we study the convergence of the proposed method in FMDPs. For FMDPs, if the state abstraction and temporally extended actions are constructed based on one state variable in each layer, then proof 1 is still valid (as shown in HEX-Q). However, the assumed condition in HEX-Q of only having one state variable for FMDPs is not a practical assumption; therefore here we evaluate the optimality of the \textit{SARM-HSTRL}'s solution for a general case. Basically, there is not a straightforward proof for convergence of methods that extract the hierarchical structure of tasks in FMDPs \citep{jonsson2006causal,mehta2011automatic}. It is proven in \citep{dean1997model} that having the stochastic substitution and reward respecting characteristics preserves optimality for reduced MDPs such as FMDPs. Thus, stochastic substitution and reward substitution can be used to prove optimality by showing that each reduced MDP has the mentioned characteristics. Next we review such characteristics of the proposed method.\\

\textbf{Definition 4:} Transaction, $\Omega$, a set of extracted trajectories, is called \textit{representative} if $\Omega$ includes all possible state action pairs that lead to the ultimate goals.

In \textit{SARM-HSTRL}, the trajectories are used instead of high-level sources of knowledge (e.g, DBNs). Since DBNs show casual relation among state variables for each action; the HRL models based on DBNs can present irrelevant states variables in state abstraction. More importantly, as we mentioned earlier, the assumption of having DBNs in advance is not practical in autonomous settings. Our proposed method solve this problem by extracting the relations among the states and state abstraction in an autonomous manner, where the trajectories are the only source of knowledge to show the effects of actions on state variables. 

\textbf{Definition 5:} A non-redundant trajectory is defined as a trajectory which is not possible to remove one or more of its state and action pairs such that the remaining sequence still leads to the goal states \citep{mehta2011automatic}. 

In \cite{mehta2011automatic}, a trajectory-task pair, $\langle \Omega_{k},T_{i} \rangle$, where $\Omega_{k}=\langle  s_{0}, a_{0}, \dots,  s_{n-1},a_{n-1},s_{n} \rangle  \subset  \Omega$, is called \textit{consistent} with \textit{$H$} if the following two conditions hold: i) subtask $T_{i}$, as an SMDP, corresponds to a node in \textit{$H$}; ii) if the observed states except the last two ones in $\Omega_{k}$ are a subset of $S_{i}$; in other words,  $ \{ s_{0},\dots, s_{n-2} \} \subseteq S_{i} $ and $\{ s_{0},\dots, s_{n-2}\} \cap G_{i} = \emptyset$. Also, $(s_{n-1}, a_{n-1})$ is an exit of $G_{i}$ of the subtask $T_{i}$. Clearly, a trajectory $\Omega_{k}$ is consistent with the extracted HST, \textit{$H$}, if $\langle \Omega_{k}, T_{0} \rangle$ is consistent with \textit{$H$}.\\

\textbf{Theorem 2:}  If each member of the set of trajectories, \textit{transactions}  $\Omega = \langle \Omega_{1}, \dots, \Omega_{m}  \rangle$, is non-redundant, \textit{SARM-HSTRL} builds a hierarchy \textit{$H$}, as every trajectory of the set is consistent with \textit{$H$}.\\

\textbf{Proof:}  Our proposed method first generates a hierarchy $H$ based on the extracted association rules of representative and non-redundant trajectories. Since a sequential ARM is utilized, it selects a sequence of states of trajectories that preserves the appeared order of them as association rules. These association rules as a whole are added to HST tree one by one and if two nodes cannot be matched, a branch of the parent node of \textit{H} is created (i.e., \textit{lines 19-22} of Algorithm 3). If a trajectory $\Omega_{k}$ is denoted by $\Omega_{k}= \langle  s_{0}, a_{0}, \dots,  s_{n-1},a_{n-1},s_{n} \rangle$, the proposed method finds the conjunction of values of $X$ that are true in $s_{n-1}$ and not before and assign that to the goal $G_{i}$ \citep{mehta2011automatic}. If there are not such values of $X$, some suffix of the sequence can be disregarded without any impacts to achieve the goal, it is a contradiction with the property of non-redundancy. As a consequence, $S_{i}$ will be the set of all states that do not satisfy $G_{i}$; thus, $\{ s_{0},\dots, s_{n-2}\}$ will satisfy the required condition to be in $S_{i}$ . 

It is noted that the trajectories can be considered as a sequence of sub-trajectories, where each of these partitioned sub-trajectories is a conjunction of values of $X$ as termination conditions of that sub-trajectory. With this, the above argument can apply to each sub-trajectory recursively \citep{mehta2008automatic}. $\Box$\\

\textbf{Definition 6:} A hierarchy is called \textit{safe} if it guarantees that ``the state variables in each task are sufficient to capture the values of any trajectory consistent with the sub-hierarchy rooted at that task node"\citep{mehta2011automatic}. In fact, the concept of \textit{safe} refers to stochastic substitution and reward respecting for sub-tasks as mentioned in \citep{dean1997model}.\\

\textbf{Theorem 3:} The hierarchical structure of tasks being extracted by \textit{SARM-HSTRL}, $H$, of a representative $\Omega$ guarantees that ``the total expected reward during each trajectory of $\Omega$ is only a function of the values of $x \in X_{i}$ in the starting state of $\Omega$" \citep{mehta2011automatic} for any trajectory-task $\langle \Omega_{j},T_{i} \rangle$ that is consistent with $H$. Also, there is just one hierarchical structure of tasks that can be extracted based on extracted exit states, subgoals, that is safe with respect to $\Omega$.\\

\textbf{Proof:} \textit{SARM-HSTRL} constructs $T_{i}= \langle X_{i},S_{i},C_{i},G_{i} \rangle $  directly based on subgoals, exit states, of several trajectories not DBNs. The subgoals are used to partition the sequence of states of trajectories, $\Omega$. The actions in any sequence of state-action pairs of each trajectory are primitive and change the values of $X_{i}$ as their resultant states are in the same partition- except exit states as termination conditions, $G_{i}$. If it changes the variables outside of current sub-task, $T_{i}$, that variable, $X_{k}$, should appear in the sequence of state-actions pairs before exit states' variables of $T_{i}$. Thus, it will be placed inside of sub-task $T_{i}$ what is a contradiction with the assumption that it can have effects on variables more than $X_{i}$ that are outside of current sub-task. In the same way, we can say that all immediate rewards in the trajectory are functions of the variables in $X_{i}$. Therefore, the summation of discounted rewards and the probability of transition in each trajectory are just related to $X_{i}$; thus, the extracted hierarchical structure of tasks is safe with respect to $\Omega$. Since the proposed method form the sub-tasks of subgoals all in once; thus, if there is another hierarchy, $H^{'}$, as it is consistent with $\Omega$, it will violate the \textit{safe} characteristic with respect to $\Omega$. This completes the prove that the extracted hierarchy by \textit{SARM-HSTR}L leads to the hierarchical optimal policy.$\Box$\\

\textbf{Theorem 4:}  The extracted hierarchical structure of tasks using \textit{SARM-HSTRL} provides the most efficient, reliable, and compact hierarchical structure considering the representative and non-redundant set of trajectories, $\Omega$, when the problem is sparse in both of MDPs and FMDPs. Efficiency is measured by the probability of usage and the resultant performance. Reliability is a function of the accuracy for the certainty of occurrence of next sub-tasks depending on which sub-tasks have been done so far. Resultant performance captures the compactness concept and is defined as how much the extracted structure abstract action space. Thus, efficient subtasks are considered as subtasks that summarized the longest frequent sequence of actions in  temporally extended actions.\\ 

\textbf{Proof:} 
Sub-tasks are extracted based on support and confidence measures in the form of association rules as the most efficient and reliable sequence of subtasks, exit states, among representative and non-redundant trajectories, $\Omega$. The \textit{support} measure checks the ratio of witnessing all possible subtasks to all observations in $\Omega$. Thus, it finds the subtasks that happen with the highest probability related to other ones. In other word, these subtasks are the best summarization, longest and most frequent, of what happened in past. Reliability implies providing the highest accuracy of predicting next sub-tasks based on summarization of several trajectories and what have been done so far. The \textit{confidence} measure evaluates every possible sequence  by constructing a tree considering all possible eligible sequences among several trajectories. It preserves their sequences and compacts the extracted subtasks in form of sequential association rules by matching and mapping them from the last task to the first ones. Also, it can be said the required size for value function table is a function of the depth, $l$, and branch, $d$, of the tree. The branch of the tree is the number of sequence of subtasks as they cannot be matched to current nodes of the tree. The depth of the tree is the number of subtasks that in the worst case equals the length of trajectory when they are not subparts of each other. Thus, the space complexity of value function tables of the hierarchy is $O(ld)$. $\Box$ \\

\subsection {Relative Advantages of \textit{SARM-HSTRL}} 

In this section, a summary of key advantages of the proposed \textit{SARM-HSTRL} related to other methods is provided.

One key contribution of this method is that despite other methods in the literature that are restricted to only MDPs or FMDPs, \textit{SARM-HSTRL} can be applied in both MDPs and FMDPs since it does not need an advance knowledge such as the state transitions, or some knowledge or constraints about the size of abstraction or reversible state transitions. Our proposed method works from scratch based on trajectories and without the need for the state transition graph or DBNs, and considers both topological and value intrinsic relationships and structures in trajectories to extract hierarchical structure of tasks. 

The proposed \textit{SARM-HSTRL} can also outperform the HI-MAT algorithm in the sense that HI-MAT only works on a single successful trajectory, while in many RL settings, there are several optimal or near-optimal trajectories that cannot be represented in HI-MAT, unless it is generalized by using another function (i.e., \textit{action generalization}). However, our proposed method does not require a single, carefully formed, trajectory, and it can efficiently handle the funnel property of subtasks, while HI-MAT cannot be generalized from many different starting places in a few terminal states (i.e., it does not have the \textit{funnel} property \citep{mehta2008automatic}).

In last, the proposed \textit{SARM-HSTRL} method can be easily scaled up to high dimensional discrete action space and even continuous action space as it considers all paths together at once. The complexity of \textit{SARM-HSTRL} is a function of complexity of FP-growth algorithm as its main component to extract the associate rules, which is proven to be very practical in terms of time complexity for real usages \citep{kosters2003complexity, tan2006introduction}.

\section{Time Complexity}
In this section, we discuss the time complexity of the proposed \textit{SARM-HSTRL}. Associate rule mining method has a considerably better performance compared to conventional  correlation extraction methods such as mutual information or statistical hypothesis testing, since they are often not able to precisely extract the intrinsic correlations among these variables \citep{tan2006introduction}. However, ARM improves upon such simple methods by using a combination of two key measures of \textit{support} and \textit{confidence} in a proven efficient extraction strategy to obtain and evaluate the most efficient and reliable relationships among the variables in datasets. 
As shown by \cite{ tan2006introduction}, for a dataset with $d$ number of items, the number of rules can be calculated as $R$ = $3^{d}$ - $2^{d+1}$ $+$ $1$. Therefore, it is impractical to enumerate all possible rules in large datasets in a naive manner.
In Rule Generation, each frequent $k$-itemset has $2^{k}-2$ rules, where $k$ is the number of items of the corresponding itemset \citep{tan2006introduction}.

As mentioned in \cite{tan2006introduction}, ``the size of a FP-tree typically is smaller than the size of the uncompressed data,'' and in the worst-case scenario, the size of a FP-tree is effectively equal to the size of the data. The performance of the FP-growth algorithm is related to the compaction factor of the trajectories and the value of \textit{minsup}. In the worst-case scenario, the support values of all combination of items are bigger than \textit{minsup}, and $2^{d+1}$ itemsets will be generated, where $d$ is the number of items. However, \textit{SARM-HSTRL} is looking for sub-goals, and the number of sub-goals in an RL task is much less than the size of state space. Thus, using the FP-growth algorithm is efficient and practical in \textit{SARM-HSTRL} when the state space is large, and the number of sub-goals is relatively low. Such sparsity is a very common assumption in HRL methods \citep{jonsson2006causal,mehta2011automatic}.

As mentioned above, \textit{SARM-HSTRL} by using FP-growth algorithm method provides a promising solution in practical applications where the state space is large and sparse. If the state space is small, or the successful trajectories have many similarities to each other, many states will be visited frequently, and hence detected by ARM as sub-goals. Clearly, the concept of sub-goals becomes meaningless in such conditions. Another possible scenario to consider is when the adjacent states around the sub-goals are visited frequently. For both these conditions, one efficient solution is to cluster the adjacent sub-goals as one entity and create one corresponding temporally extended action for that entity. $t$, order of occurrence, for each state in each trajectory is already stored by \textit{SARM-HSTRL} as they are used in HST for possible orderings of sub-goals. They can be also used to find the close sub-goals for clustering purposes.

\begin{figure*}
	\centering
	\subfigure[]{\label{fig:maze_action_1_z}\includegraphics[width=70mm, height=1.3in]{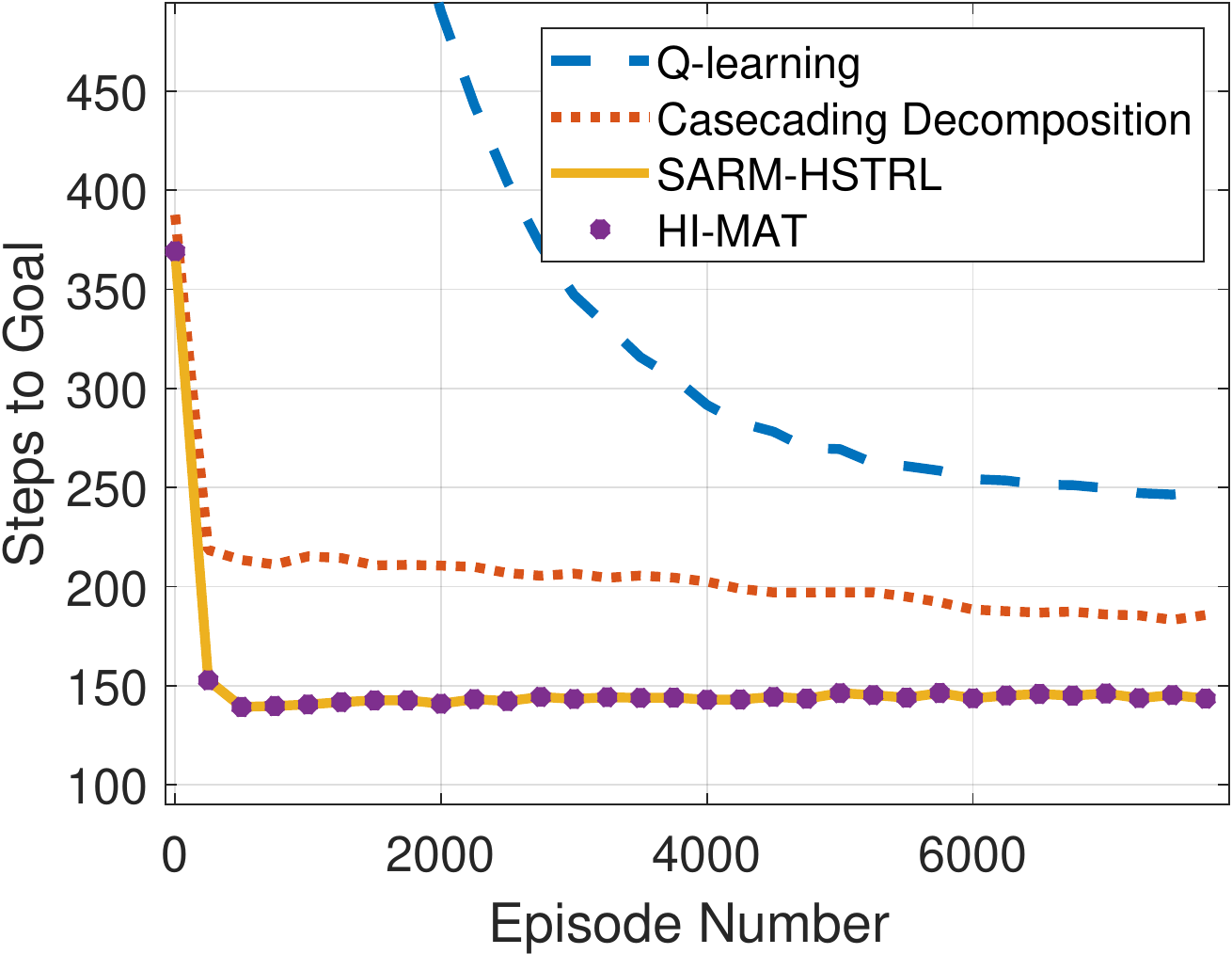}}
	\subfigure[]{\label{fig:maze_reward_1_z}\includegraphics[width=75mm, height=1.3in]{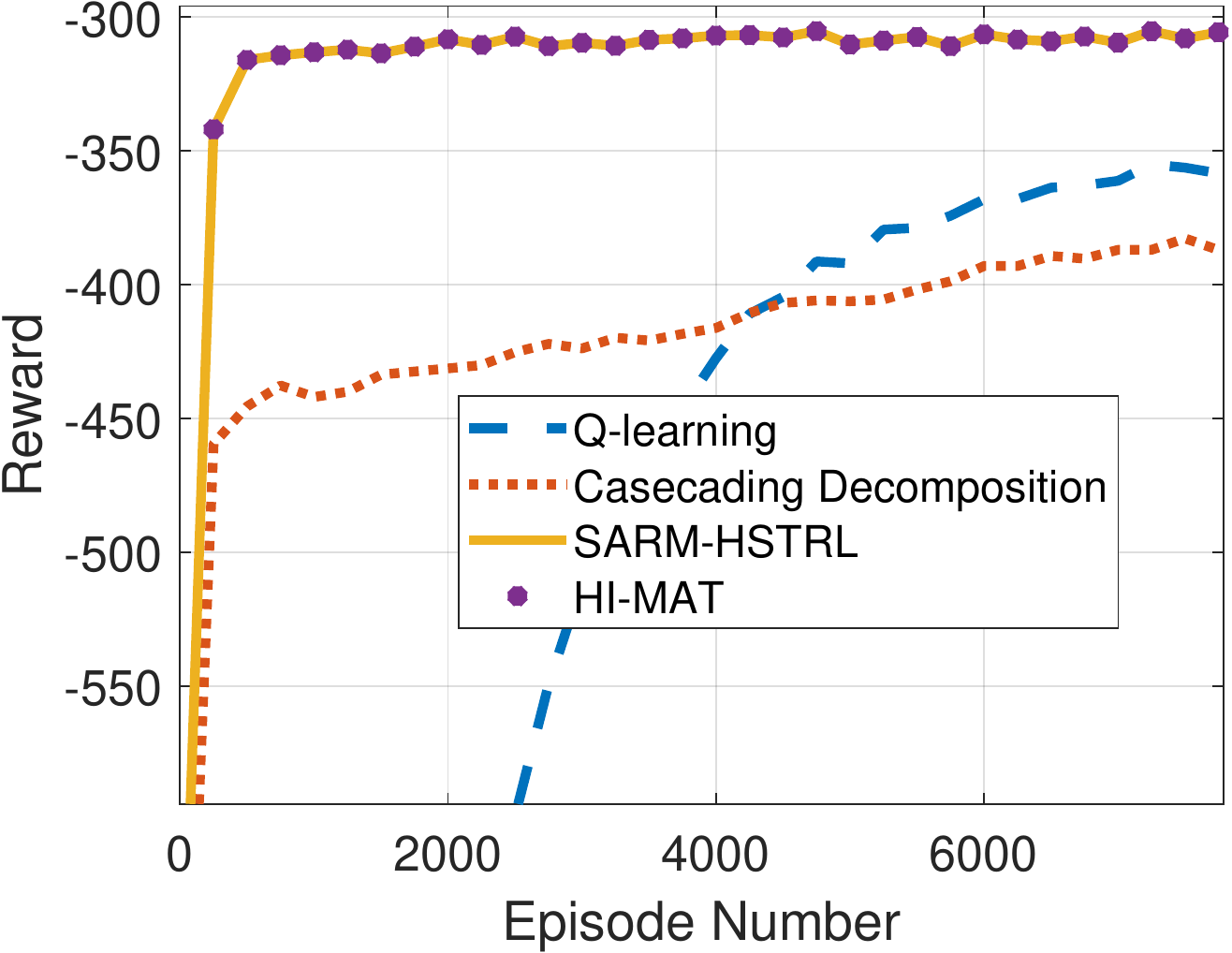}}	
	\subfigure[]{\label{fig:action2zlabel}\includegraphics[width=70mm, height=1.3in]{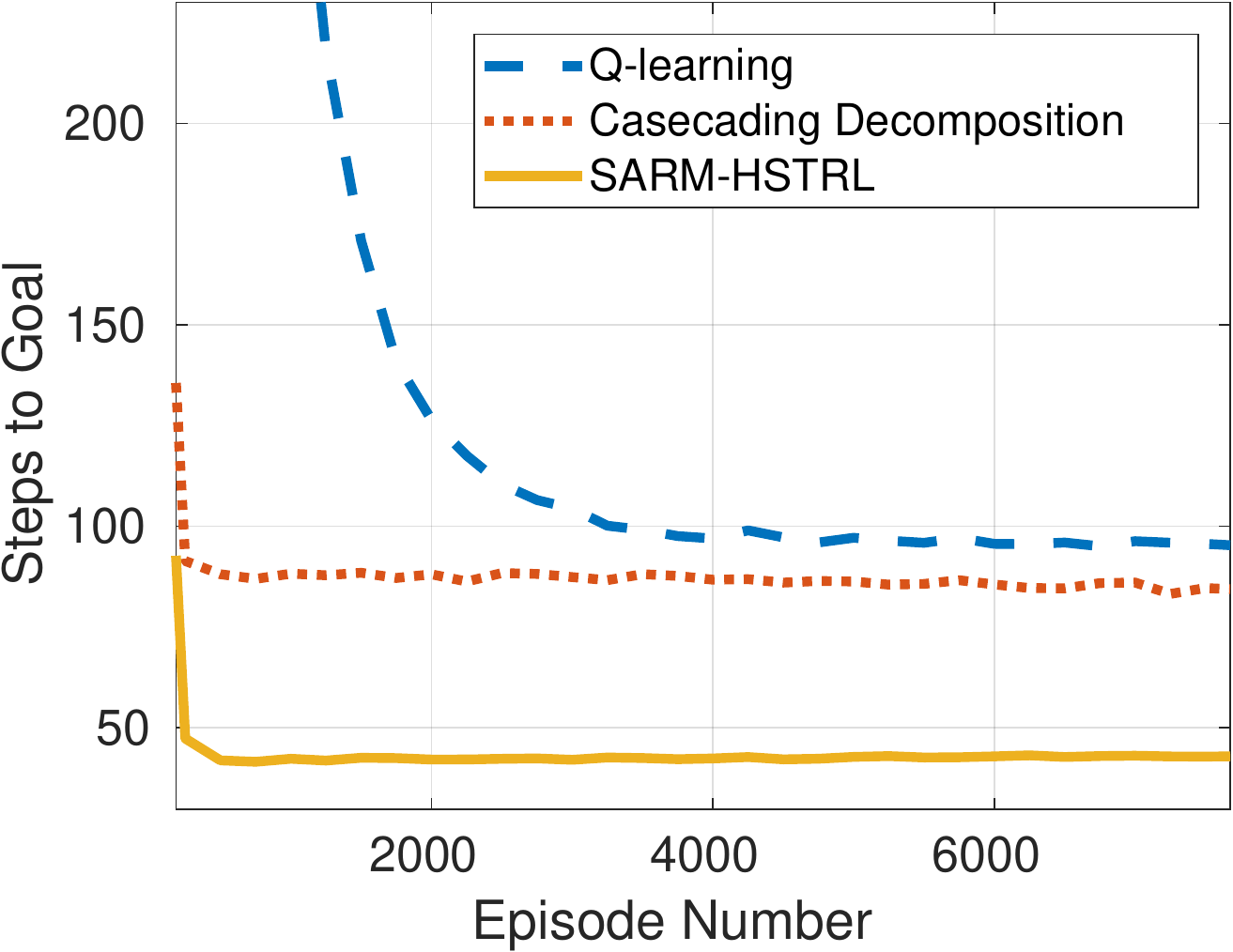}}
    	\subfigure[]{\label{fig:reward_2_z}\includegraphics[width=75mm, height=1.3in]{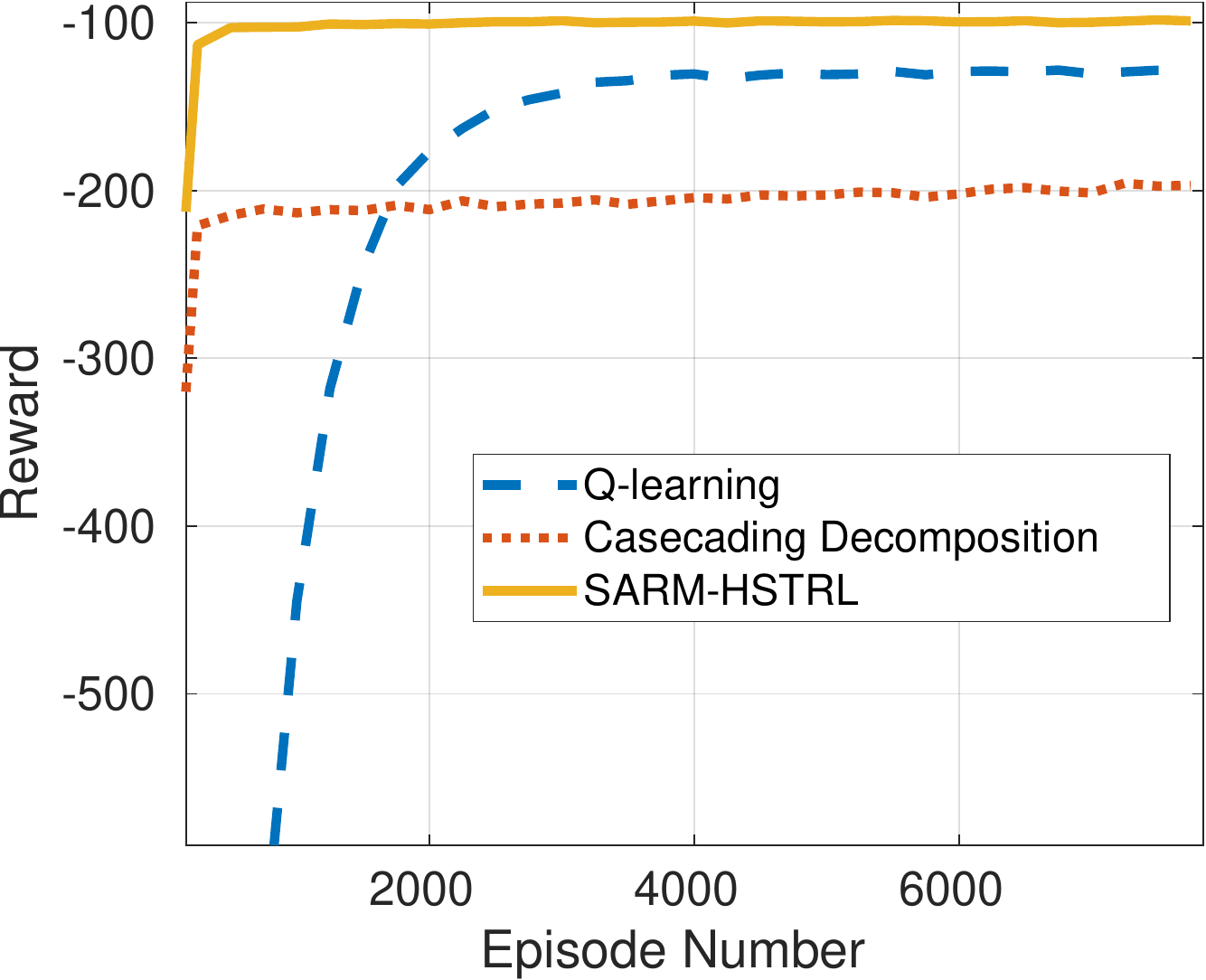}}
        \vspace{-10pt}
\caption{ \small {Performance comparison of \textit{SARM-HSTRL} with Q-learning, HI-MAT, and Cascading Decomposition in experiment 1 (Fig. \ref{fig:HST_test1_hi}) and experiment 2 (Fig. \ref{fig:HST_test2_hi}) of the first testbed described in Figure \ref{fig:testbed1}.  Since experiment 2 including multiple successful trajectories, HI-MAT cannot be implemented. HI-MAT can only work with one successful trajectory that interprets the tasks.(a) Represents the number of steps along episodes in experiment 1.  (b) Comparison of receiving rewards along episodes in experiment 1. (c) Represents the number of steps along episodes in experiment 2. (d) Comparison of received rewards along episodes in experiment 2.}}
\vspace{-10pt}
\end{figure*}

\section{Experimental Results} \label{con}


In this section, several experimental results are presented to evaluate the performance of \textit{SARM-HSTRL} on four different testbeds. In the first two experiments, the agent has 5 actions, \textit{press-key} and 4 movement primitive actions. The \textit{press-key} does not change the place of the agent. The agent can move with its primitive actions in four directions: \textit{up, right, down, left}. If there is a wall in the way, the agent stays in its current state. In all of the experiments, if the agent does the \textit{press-key} action, it will receive a reward of $0$ in the sub-goal places and a reward of $-10$ in other states. The reward of other actions is $-1$. The agent movement with probability $0.8$ is according to an intended action and is randomly in one of the directions with probability $0.2$. The discount factor is set to $\gamma$= $0.9$.

In constructing the HST, $10$ start and goal places are chosen randomly. A goal state is defined as an important, task-specific state that ends an episode once visited. A start state, $s_{0}$, is a state from which an agent begins an episode. For each of them, the agent starts the learning using a common learning mechanism such as Q-learning; the learning is finished after 5000 episodes. They are ordered based on the accumulated reward, and the best five ones are selected. They are given to the \textit{SARM-HSTRL} and the HST produces a hierarchal structure of tasks based on the whole length of transactions. Now, the subtasks are formed for the agent and the HST helps the agent to choose their phase of learning. 
If they are expanded as primitive actions, the number of steps to reach a goal is equal to the number of action selection calls. 

The performance of \textit{SARM-HSTRL} in HRL is evaluated in Figure \ref{fig:testbed1} for two different hierarchical structure of tasks, experiment 1 and experiment 2. In these figures, for the sake of comparison between Q-learning, Cascading Decomposition \citep{chiu2010automatic}, HI-MAT \citep{mehta2008automatic}  and \textit{SARM-HSTRL}, 10 runs are considered where in each of them, a start state and a goal state are chosen randomly. The maximum number of actions for each episode is 4000, and the total number of episodes is 8000.  \textit{SARM-HSTRL} is compared to Cascading Decomposition as a representative approach in MDPs, which as discussed in \citep{chiu2010automatic,ghazanfari2016extracting}, is the latest and considerable improvement for methods proposed in  \citep{csimcsek2009skill,mannor2004dynamic,stolle2004automated}. As seen in this figure, the proposed \textit{SARM-HSTRL} method results in a hierarchical optimum policy task structure, as does HI-MAT, while our method does not rely on any prior knowledge (e.g.  DBNs). It has been proven in \citep{mehta2008automatic} that HI-MAT leads to better results compared to VISA, and this concludes that \textit{SARM-HSTRL} outperforms the VISA method too. It is worth mentioning that HI-MAT cannot be implemented in experiment 2 including multiple successful trajectories, as it can only work with one successful trajectory that interprets the tasks.

In experiment 1, Figure \ref{fig:HST_test1_hi}, the task hierarchy has 7 levels -- it has $(484\times(7+1))=3872$ states. If the agent enters in sub-goals states in the following order $1,2,3,4,5,6 $ and $7$ and does the \textit{press-key} action in each of them, and then enters in the goal state of the run and performs the \textit{press-key} action again, the agent receives a reward of $+10$, and the episode will be finished. The value of \textit{minsup} is 0.9 and the value of \textit{minconf} is 0.9. 


In experiment 2, Figure \ref{fig:HST_test2_hi}, the task hierarchy has 4 levels, but with a more complicated structure- it  has $(484\times(4+1))=2420$ states. If the agent enters in one of the sub-goals states from the leaves of tree $1$ or $2$  or $3$, then enters in one of their parent $4$ or $5$, then in $6$ and $7$ in order and does the \textit{press-key} action in each of them, and finally enters in the goal state of the run and performs the \textit{press-key} action, the agent receives a reward of $+10$ and the episode will be finished. The value of \textit{minsup} is 0.3 and the value of \textit{minconf} is 0.9.
\begin{figure*}
	\centering
	\subfigure[]{\label{fig:testbed1_light}\includegraphics[width=65mm, height=1in]{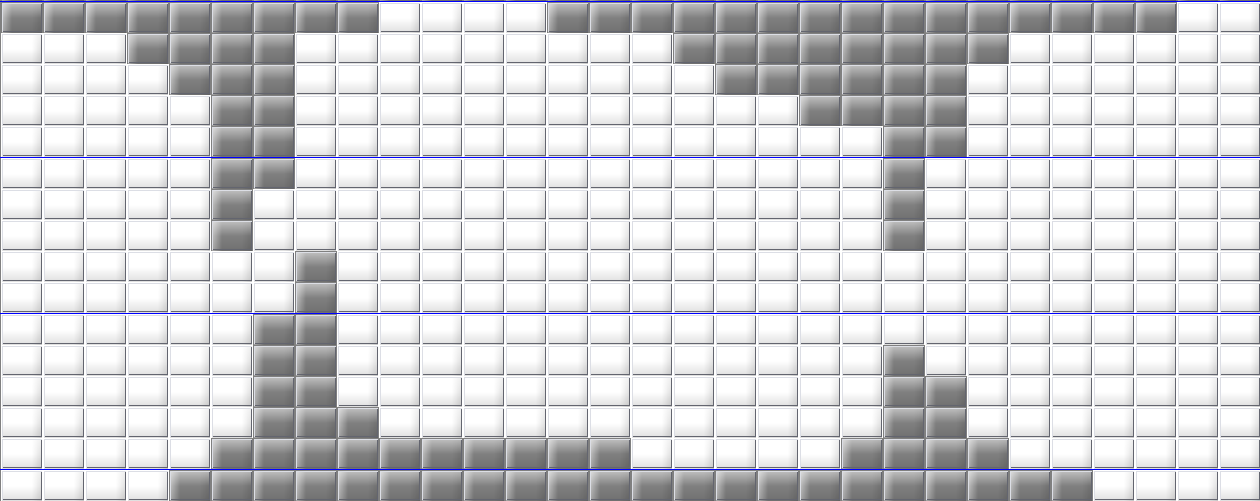}}
	\subfigure[]{\label{fig:testbed_light}\includegraphics[width=70mm, height=1in]{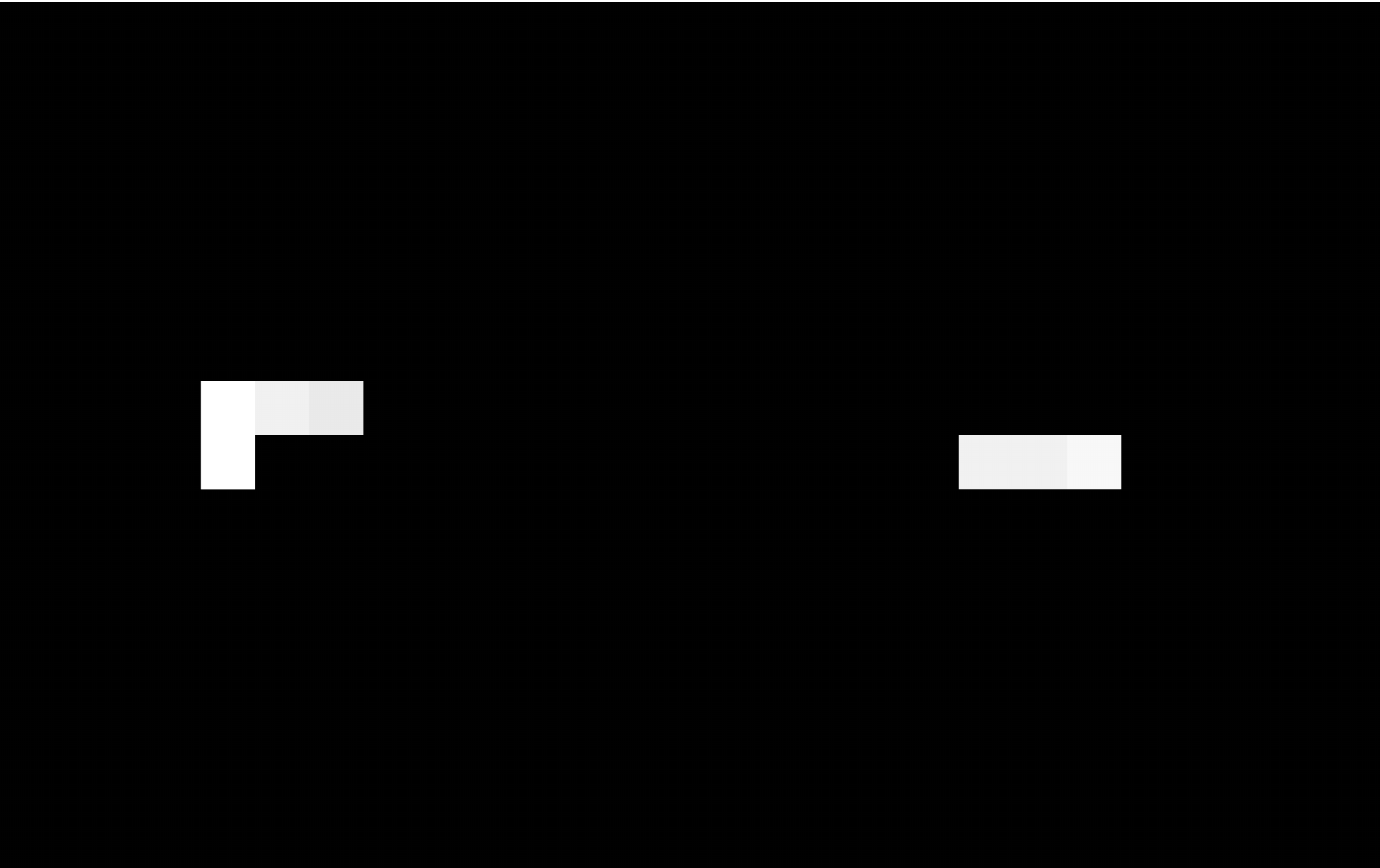}}	
	\subfigure[]{\label{fig:Taxi_d}\includegraphics[width=65mm, height=1in]{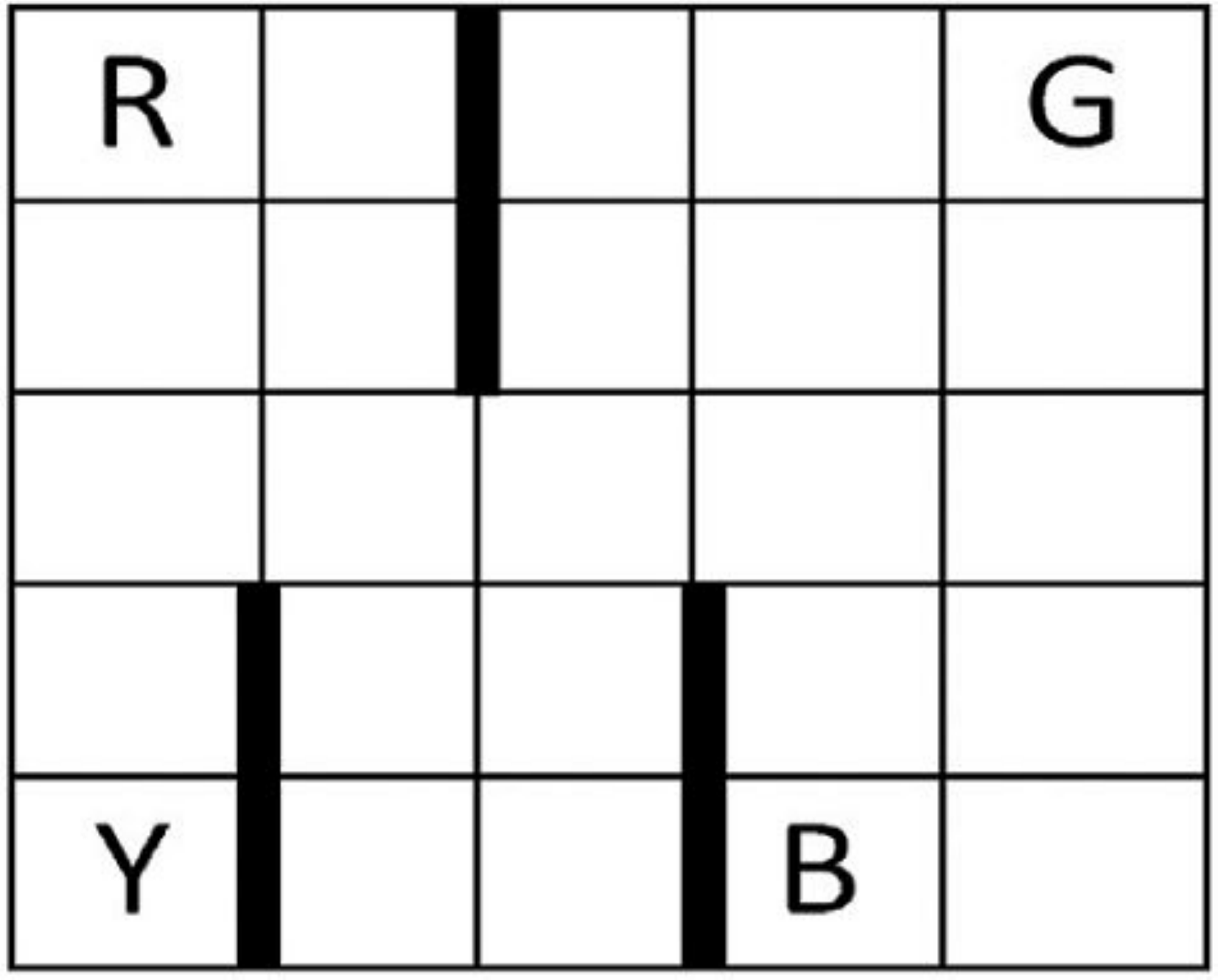}}
	\subfigure[]{\label{fig:Taxi}\includegraphics[width=70mm, height=1in]{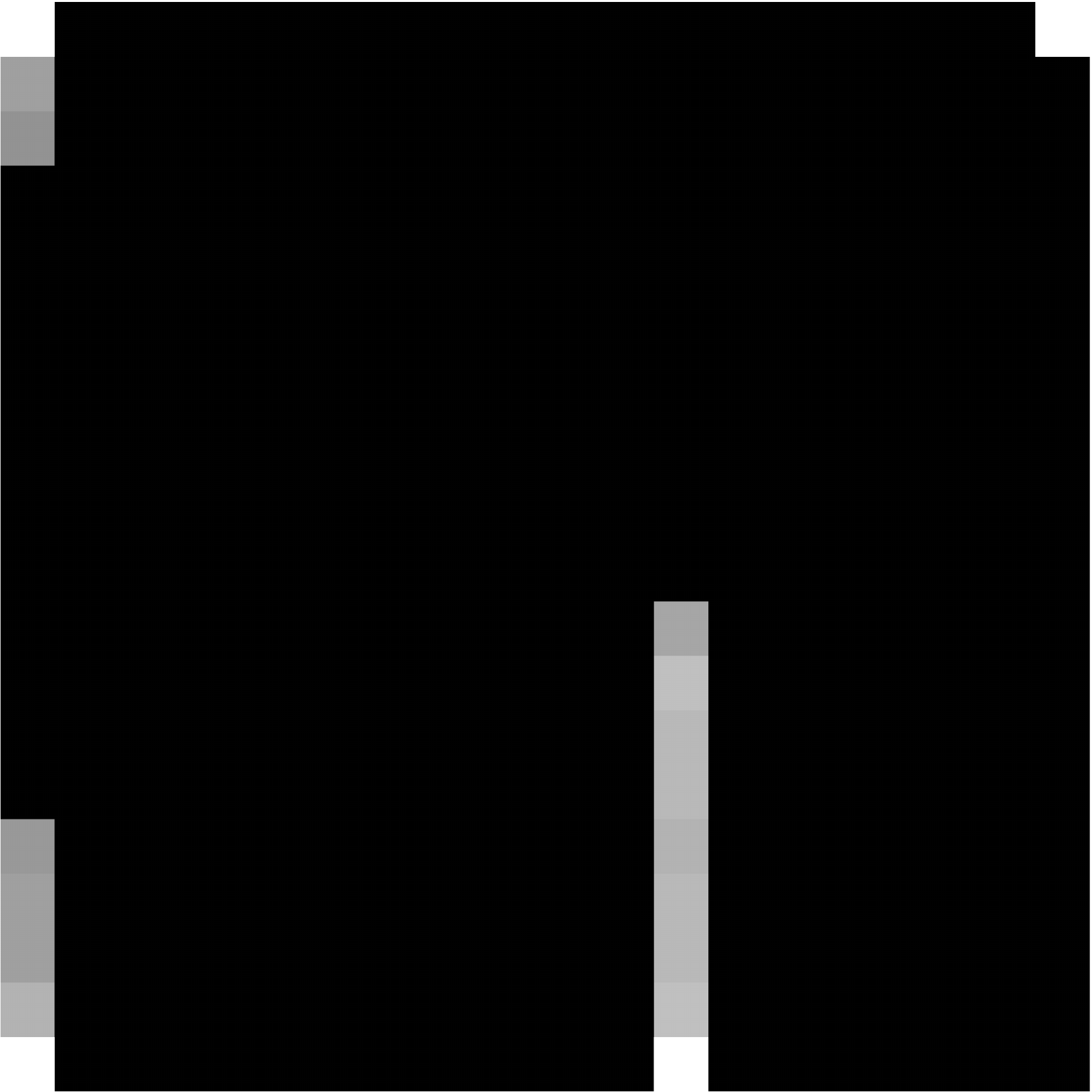}}	
    \vspace{-10pt}
\caption{ \small{(a) A maze world. (b) The frequency of visiting detected subgoals by \textit{SARM-HSTRL} in transitions. (c)  Taxi driver problem as an example in FMDPs. (d) The frequency of visiting the detected subgoals by \textit{SARM-HSTRL} in transitions in a 4 times scale in places' dimensions of Taxi driver problem, 16 times larger state space. The states near to wall states of three places are more probable to visit because of they  experience less influence of the stochastic rate and the place of pick up places. Also, the four places as the \textit{SARM-HSTRL} are detected correctly that have the most observing, the brightest ones. }}
\vspace{-10pt}
\label{fig:light_gray}
\end{figure*}
There is a significant difference in speed of learning between the proposed method with Cascading Decomposition and Q-learning as shown in Figures \ref{fig:maze_reward_1_z} and \ref{fig:reward_2_z}). The most important attribute of SMDP framework is using temporally extended actions to decrease the number of steps. As it is shown in HRL in Figures \ref{fig:maze_action_1_z}  and \ref{fig:action2zlabel}, the temporally extended actions considerably decrease the number of steps. p-values have been calculated between the proposed method with Q-learning and Cascading Decomposition in each diagram by using the t-test for $\alpha$ = $0.01$; the significant change is validated -- p-values are much smaller than $1\times{10^{-5}}$. 

In experiment 3, the accuracy of \textit{SARM-HSTRL} is evaluated thorough all possible subgoals, where 10 random states for start and goal states are selected. The minsup and minconf are set to 0.6 and 0.9, respectively for Figure \ref{fig:testbed1_light}. The agent has four actions \textit{up, right, down, and left}. Both the stochastic rate and learning rates are 0.1, and the discount factor and the e-greedy are the same as the previous experiments. The agent receives a reward of zero for each action, unless enters to the goal state where it receives 10. The number of trial is 500 for each pair of start and goal states that 5 of the best trajectories are used. As it can be seen in Figure \ref{fig:testbed_light}, \textit{SARM-HSTRL} detects the subgoals properly. \textit{SARM-HSTRL} with the given threshold did not consider all the possible subgoals in the right side of Figure \ref{fig:testbed1_light} since the middle ones are placed in better policies, they can reach the possible goals with more probability and less actions.    

In experiment 4, we aim to show the accuracy of \textit{SARM-HSTRL} in FMDPs, using Taxi driver problem as a known testbed (Fig. \ref{fig:Taxi_d}). We scale up both place dimensions of Taxi driver problem for 4 times to reach $20\times{20}$. The taxi domain is composed of a $5  \times  5$ grid world, a taxi, and a passenger, where the taxi starts from a random place and \textit{pick-up} the passenger from one of those places (\textit{B, G, R, and Y}) and \textit{put-down} the passenger in one of these places. The place of \textit{pick-up} and \textit{put-down} are chosen randomly. The taxi has six primitive actions, \textit{north, south, east, west, pick-up, and put-down}. The agent receives a reward $-1$ for movement actions, a reward $-10$ for wrongly doing the action \textit{pick-up or put-down}, and a reward $+20$ for successfully completing the mission. Each action succeeds in its job with the probability of $0.8$ in each state and it has a random effect in that state with the probability of $0.2$.  The number of trials is $2000$, and $16$ random start and goal states to capture all possible combinations of \textit{pick-up} and \textit{put-down}. The maximum number of action is $1000$ in each trial.  \textit{minsup} is set to 0.0625 and \textit{minconf} is set to 0.7. The \textit{minsup} value is selected as 0.0625  noting that there are 16 combinations for \textit{pick-up}, and \textit{put-down}. Discount factor, $\epsilon - greedy$, and learning rate have been initialized similar to experiment 1.  As is shown in \ref{fig:Taxi}, the number of observing detected subgoals for \textit{pick-up} is correct (the brightest ones). Also, some states in the paths to subgoals are visited more frequently, therefore  are being detected as the subgoals. For example, when these states are in the optimal paths of several subgoals, or they are adjacent to the wall states,  they will be visited more  because of the stochastic rate. They can be easily pruned by considering the sequence and their adjacency to states with biggest support. There is another way in such condition, where the adjacent extracted sub-goal states can be considered as a cluster to define just one temporally extended actions for them.

\section{Conclusion} \label{con}
\noindent A HRL method called \textit{SARM-HSTRL} is proposed to autonomously extract a task hierarchy for RL by utilizing a sequential associate role mining approach, where multiple subgoals are extracted as frequently visited states from successful trajectories in the form of association rules. These subgoals are used to define exits as termination conditions to form temporal and state abstractions. Despite the majority of the previously proposed HRL methods (e.g., HI-MAT and VISA) that rely on DBNs model to use prior knowledge about the effects of actions on state variables, our proposed method independently extracts the relations among states and state abstraction. Moreover, since DBNs show the causal relations among the state variables for each action, it can determine irrelevant states variables for state abstraction. However, the proposed method only extracts the relevant correlations. The convergence of the proposed method to a hierarchical optimal solution is proven for both MDPs and FMDPs. The experimental results show a considerable improvement in the speed and quality of the learning process for the analyzed  experiments. It is expected that the extracted hierarchical structure in the form of sub-tasks provides a supplementary, and a more robust and higher level of knowledge to be transfered among the sub-tasks rather than sharing value functions, which are highly sensitive to the type and the amount of similarity between the source and target domains. Therefore, the decomposed structure of tasks based on \textit{SARM-HSTRL} provides an abstraction that an agent can reuse, generalize, and transfer to new domains.

\bibliography{reff}
\bibliographystyle{aaai}

\clearpage

\section{Supplementary Material}

\subsection{Background}
In this section, a brief introduction on Markov decision process (MDP) and factored MDP (FMDP) and the corresponding notations in the main paper is provided.\\

\noindent \textbf{MDPs and FMDPs:} \noindent RL tasks are typically defined in a Markov Decision Process (MDP) framework as a $5-tuple$: $\langle S, A, P, R, \gamma  \rangle$. In this paper, we focus on finite MDPs, where 
$S$=$\{s_{1}, \dots, s_{n}\}$ is a finite set of states, 
$A$=$\{a_{1}, \dots, a_{m}\}$ is a finite set of primitive actions, 
$P: S \times A \times S \rightarrow [0,1]$ is a one-step probabilistic state transition function, 
$R:S\times A \rightarrow \mathbb{R}$ is a reward function, 
and $\gamma \in (0,1]$ denotes the discount rate. 
The agent's goal is to find a policy (a mapping from states to actions), $\Pi:S \times A \rightarrow [0,1]$ that maximizes the accumulated discounted reward $R=\sum_{i=0}^{T} \gamma^i r_i$, for each state in $S$. FMDPs are known as an extension of MDPs that contain structured representation of problems, where $T$ and $R$ are represented in a compact way. In factored MDPs, the states are described by a set of state variables. To have a unified definition for both MDPs and FMDPs, each state in a MDP can be described by a random variable $X$ which contains one variable $X_1$,  $X$= $(X_{1})$, that takes different values. 
In FMDPs, $X$ is a multivariate random variable, $X=( X_{1},X_{2},\dots,X_{n})$. Each state $x$ is an instantiation of $X$, and it can be shown as a vector of $(x_{1},x_{2},\dots,x_{n})$ such that $\forall i$ $x_{i} \in Dom(X_{i})$, in which $DOM(X)= \langle D_{1},D_{2},\dots, D_{n}  \rangle$ refers to the set of possible values for $X$ as a multivariate variable \citep{sigaud2013markov}. 

The value of a state $s$ based on a policy $\pi$ is defined as follow: there is always at least one policy that its expected return is equal or greater than any other policies for all states \citep{sutton1998reinforcement}. Such policy or policies are known as optimal policies and shown with $\pi^{*}$. Hence, the corresponding state-value function, $V$, and action-value function, $Q$, are optimal and shown as follows: $V^{*}(s)=\max_{\pi} V_{\pi}(s)$ for all $s\in S$, and $Q^{*}(s,a)=\max_{\pi} Q_{\pi}(s,a)$ for all $s\in S$ and for $a \in A(S)$, respectively.

\subsection{Definitions}
Here, we present the definitions of terms used throughout the main paper. Also, all of the following terms has been further explained with some examples in ``An Example of \textit{SARM-HSTRL}" section.\\

\textbf{Exit}: Transitions are considered \textit{unpredictable} when they lead to entering or leaving subgoals. The region and the boundaries among states' clusters that have unpredictable transitions are considered as \textit{exits} and defined by a state action pair $G_{i}=(s^{T_{i}}, a)$ when taking action $a$, as a primitive action, from state $s^{T_{i}}$, as a subgoal, leads to the resultant state that is a goal state to complete subtask $T_i$ \citep{hengst2003discovering}.\\

\textbf{Subtask}: A subtask, $T_{i}$, is a semi-MDP (SMDP) that is shown by $\langle X_{i},  S_{i}, G_{i}, C_{i}, \rangle$ \citep{mehta2011automatic}, where $X_{i}$ is the set of variables that their corresponding values change during performing the subtask, $S_{i}$ denotes the set of admissible states of $T_{i}$, $G_{i}$ shows the exits of corresponding subtasks as termination conditions of $T_{i}$, and $C_{i}$ is the set of child tasks of $T_{i}$. Child tasks, $C_{i}$, can be formed based on different HRL frameworks such as MAXQ or option.\\

\textbf{Successful Trajectories}: A successful trajectory is defined as a trajectory of states that leads to the goal reward \citep{mehta2008automatic}.\\

\textbf{ Abstract State and  Abstract Action}: In the task hierarchy, subtasks are considered as abstract states and their corresponding local policies as abstract actions \citep{hengst2003discovering}.\\

\textbf{ Region}: A region is a set of states that are reachable of each other ``such that any exit state in a region can be reached from those state with probability 1 \citep{hengst2003discovering}.\\

\subsection{An Example of \textit{SARM-HSTRL}}\label{sec:Example_HSTRL}

\begin{figure}
	\centering 
	\includegraphics[width=3.2in, height=8.4in]{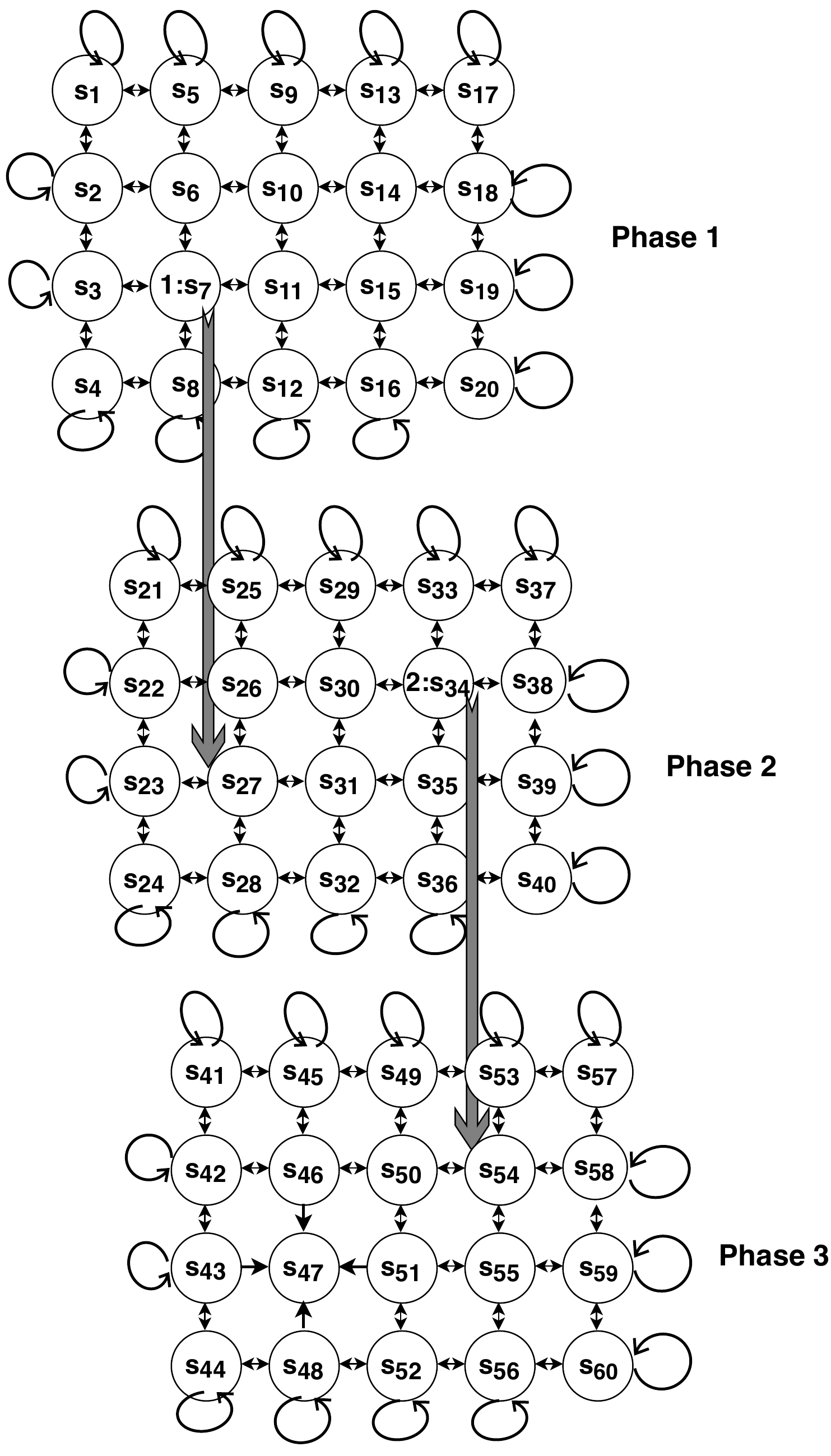}
	\caption{The example testbed: the size of the testbed is $4*5*3=60$ states.}\label{fig:exit_maze}
\end{figure}

In this section, we provide a detailed example to describe the proposed \textit{SARM-HSTRL} on a testbed described in Figure \ref{fig:exit_maze}. To define a association rule, a pair of $\langle ITEMSET,\ Transaction \rangle$ needs to be defined. \textit{ITEMSET} is equivalent with $S$, $S$=$\{s_{1}, \dots, s_{n}\}$ in which $n$ denotes the size of state space. Therefore, \textit{ITEMSET} = $\{s_{1}, \dots ,s_{60}\}$ in the following example.
\textit{Transaction} = $\{\Omega_{1}, \dots ,\Omega_{N}\}$ is the set of all transactions. As mentioned earlier, each transaction is defined as a successful trajectory. In other words, each transaction is a trajectory of states from a start state to a goal state. Since start states and goal states are chosen randomly, the first elements as start states and last elements as goal states of these trajectories can be different.

Let us define an experiment to describe the different steps and terms of the \textit{SARM-HSTRL} in the following maze depicted in Figure \ref{fig:exit_maze}. In this experiment, there are 3 phases in the system and the agent has five primitive actions: \textit{up, right, down, left, and enter}. The goal states are in the third phase. The agent starts from the first phase and can move in the second phase if the agent enters state $s_{7}$ and takes the action \textit{enter}. The third phase activates if the agent is in the second phase and enters $s_{34}$ and does the action \textit{enter}. There are 60 states, where the first four actions are movement ones (i.e., \textit{up, right, down, left}) and  action \textit{enter} can take the agent to the next phase. The device starts from a random place, and should pass through states $s_{27}$ and $s_{54}$, and go to the goal states, which are selected randomly to receive the goal reward. There is a positive reward to reach the goal state by passing these phases in the right order and a negative smaller reward for taking each action.  If we run the Q-learning method for the agent on this maze for different start and goal states, it learns policies gradually during different episodes. An episode is a trajectory of sequence of states and actions and it leads to a goal reward if it reaches the goal state that is in the third phase. Clearly, many episodes in the first of running will not lead to the goal reward. But, Q-learning learns gradually policies to reach the goal state that goes through $s_{27}$ and $s_{54}$. We consider three runs, each run corresponds to a different start and goal state, and 200 episodes of learning for each run. Among 200 episodes of each run, we select the two ones that have the biggest accumulated rewards as follows:\\

\noindent The first run: the start state is $s_{1}$ and the goal state is $s_{57}$.

$\Omega_{1}$=$\{s_{1},s_{2},s_{3},s_{7},s_{27},s_{31},s_{35},s_{34},s_{54},s_{58},s_{57}\}$.

$\Omega_{2}$=$\{s_{1},s_{5},s_{6},s_{7},s_{27},s_{31},s_{30},s_{34},s_{54},s_{53},s_{57}\}$.\\

\noindent The second run: the start state is $s_{3}$ and the goal state is $s_{59}$.

$\Omega_{3}$=$\{s_{3},s_{7},s_{27},s_{26},s_{30},s_{34},s_{54},s_{55},s_{56},s_{60},s_{59}\}$.

$\Omega_{4}$=$\{s_{3},s_{7},s_{27},s_{31},s_{35},s_{34},s_{54},s_{58},s_{59}\}$.\\

\noindent The third run: the start state is $s_{12}$ and the goal state is $s_{59}$.

$\Omega_{5}$=$\{s_{12},s_{8},s_{7},s_{27},s_{26},s_{30},s_{34},s_{54},s_{58},s_{59}\}$.

$\Omega_{6}$=$\{s_{12},s_{11},s_{7},s_{27},s_{26},s_{30},s_{31},s_{35},s_{34},s_{54},s_{58},s_{59}\}$.\\

In the above example, \textit{ITEMSET}=$\{s_{1}, \dots, s_{60}\}$, \textit{Transaction} = $\{\Omega_{1}, \dots ,\Omega_{6}\}$, and the number of transaction is 6 ($N$= 6). All of the mentioned transactions lead to the goal states of their runs, it means they are successful trajectories. To calculate the association rules in the \textit{ITEMSET}, we need to calculate the support and confidence factors for each possibility of extracted association rules in form of $A \rightarrow B$ that is performed using FP-growth algorithm noting its efficient performance. As mentioned, support and confidence are calculated as follow:

\[
\textit{support} (A \rightarrow B) = \frac{\sigma(A \cup B)}{N} \]

\[\textit{confidence} (A \rightarrow B) = \frac{\sigma(A \cup B)}{\sigma (A)}
\]
For instance, if $A$=${s_{1}}$ and $B$=${s_{58}}$, \textit{support} ($s_{1}$ $\rightarrow$ $s_{58})= \frac{1}{6}$. Since $s_{1}$ and $s_{58}$ are only simultaneously observed in $\Omega_{1}$; thus, $\sigma$ ($s_{1}$ $\cup$ $s_{58}$) = 1. The \textit{confidence} ($s_{1}$ $\rightarrow$ $s_{58}$)= $\frac{1}{1}$. 
Here, we use the sequential ARM (\textit{SARM}) technique rather than conventional ARM technique due to its capability to consider the order of items in addition to their occurrence frequency. In this case,  the \textit{confidence} ($s_{58}$ $\rightarrow$ $s_{1}$)= 0. 

\begin{figure}
	\centering	
	\includegraphics[width=15mm,height=2.0in]{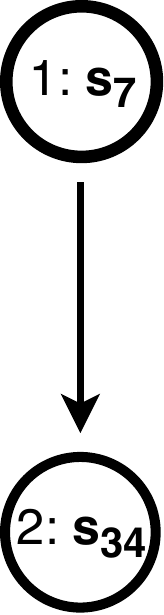}
	\caption{The extracted task hierarchy of the Figure \ref{fig:exit_maze}. An example of a sequential association rule mining and HST-construction of one rule.}\label{fig:SARM}
\end{figure}

As another example, if we consider $A$=${s_{7}}$ and $B$=${s_{34}}$,  \textit{support} ($s_{7}$ $\rightarrow$ $s_{34})= \frac{6}{6}$. Since $s_{7}$ and $s_{34}$ are visited in all $\{\Omega_{1}, \dots ,\Omega_{6}\}$; thus, $\sigma$ ($s_{7}$ $\cup$ $s_{34}$) =6. The \textit{confidence} ($s_{7}$ $\rightarrow$ $s_{34})= \frac{1}{1}$. 
If we set $minsup$=0.9 and $minconf$=0.9, $s_{7}$, $s_{27}$, $s_{34}$, and $s_{54}$ validates. Since $s_{7}$ and $s_{27}$ are consecutive by one action, \textit{enter}, the exit is defined as ($s_{7}$, \textit{enter}). In the same way for $s_{34}$ and $s_{54}$, ($s_{34}$, \textit{enter}) is considered as the exit. Therefore, only states $s_{7}$ and $s_{34}$ are used for \textit{HST-construction} method and the result structure is presented in Figure \ref{fig:SARM}.

The proposed \textit{SARM-HSTRL} method extracts the exits $G_{0}=(s_{7}, enter)$ and $G_{1}=(s_{34}, enter)$ to form subtasks $T_{0}$ and $T_{1}$. The edge between subtasks $T_{0}$ and $T_{1}$, denoted by $E$, in the extracted graph shows the relation between these subtasks. Based on the extracted exits, the transactions are partitioned into three regions as follows:  $\{s_{1}, \dots, s_{20}\}$, $\{s_{21}, \dots, s_{40}\}$, and $\{s_{41}, \dots, s_{60}\}$. $T_{0}$ as the subtask is formed of $S_{0}$=$\{s_{1}, \dots, s_{20}\}$ and $G_{0}=(s_{7}, enter)$. It does not have any child. $T_{1}$ as the subtask is formed of $S_{1}$=$\{s_{21}, \dots, s_{40}\}$, $G_{1}=(s_{34}, enter)$, and its child is $C_{1}$=$T_{0}$. $T_{0}$ and $T_{1}$ nodes can be considered as \textit{abstract states} and their corresponding policies as \textit{abstract actions}. The policy of subtask $T_{0}$ is a local policy for $S_{0}$ that leads reaching to $s_{27}$ by $G_{0}$. The policies of subtasks can be obtained of the episodes. The sequences of states of episodes are used as transactions to extract exits, the corresponding actions that lead to exits are used to extract the policies.

\begin{figure}
	\centering 
	\includegraphics[width=62mm,height=2.4in]{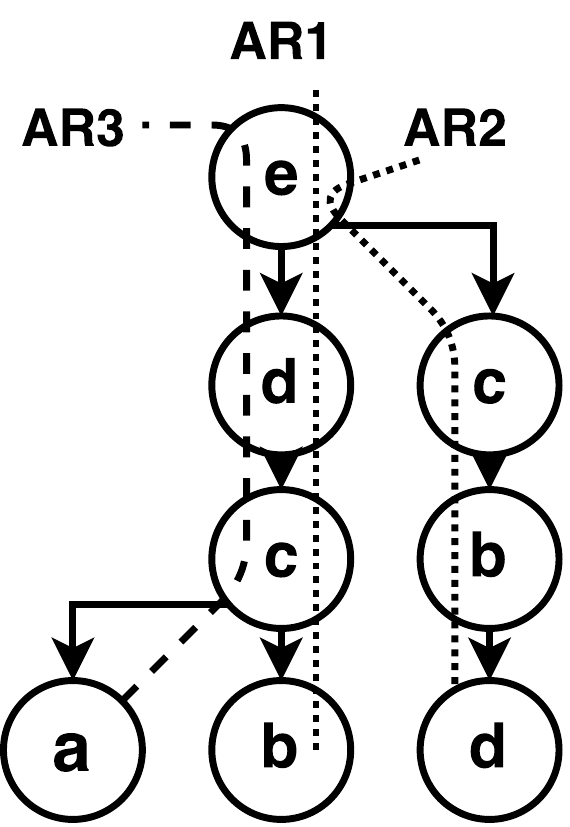}
	\caption{An example of a HST-construction of several rules.}\label{fig:HSTexample1}
\end{figure}

\subsection{An Example of HST Construction of Several Rules }
As an example, let us consider $AR_{1} = bcde$, $AR_{2} = dbce$, $AR_{3}= acde$ (see Figure \ref{fig:HSTexample1}). The proposed algorithm, first constructs a tree with the reverse of $AR_{1}$, creating one branch with values $edcb$. Then, the reverse of $AR_{2} $ is added to the tree, making a new branch from $c$ since $AR_{2,2}=c$  cannot be matched in the tree from that point. Thus, a new branch from $e$ is created to contain the remaining values of $AR_{2}$. Finally, the reverse of $AR_{3}$ is added to the tree, where a mismatch happens in $AR_{1,4}$ that results in a new branch created at node $c$.\\

\bibliography{reff}
\bibliographystyle{aaai}
\end{document}